\documentclass[12pt]{article}

\usepackage[T1]{fontenc}
\usepackage{graphicx}
\usepackage{caption}
\usepackage[numbers]{natbib}  
\usepackage{hyperref}
\usepackage{float}
\usepackage[a4paper, margin=1in]{geometry}

\title{Effects of Swarm Size Variability on Operator Workload}

\author{
  William Hunt\textsuperscript{1} \and
  Aleksandra Landowska\textsuperscript{2} \and
  Horia A. Maior\textsuperscript{2} \and
  Sarvapali D. Ramchurn\textsuperscript{1} \and
  Mohammad Soorati\textsuperscript{1}
}

\date{
  \textsuperscript{1}University of Southampton \quad
  \textsuperscript{2}University of Nottingham \\[0.5em]
  \small Preprint. Accepted at AHFE, 2026.
}

\begin{document}
\maketitle

\begin{abstract}
Real-world deployments of human--swarm teams depend on balancing operator workload to leverage human strengths without inducing overload. A key challenge is that swarm size is often dynamic: robots may join or leave the mission due to failures or redeployment, causing abrupt workload fluctuations. Understanding how such changes affect human workload and performance is critical for robust human--swarm interaction design. This paper investigates how the magnitude and direction of changes in swarm size influence operator workload. Drawing on the concept of workload history, we test three hypotheses: (1) workload remains elevated following decreases in swarm size, (2) small increases are more manageable than large jumps, and (3) sufficiently large changes override these effects by inducing a cognitive reset. We conducted two studies (N = 34) using a monitoring task with simulated drone swarms of varying sizes. By varying the swarm size between episodes, we measured perceived workload relative to swarm size changes. Results show that objective performance is largely unaffected by small changes in swarm size, while subjective workload is sensitive to both change direction and magnitude. Small increases preserve lower workload, whereas small decreases leave workload elevated, indicating workload residue; large changes in either direction attenuate these effects, suggesting a reset response. These findings offer actionable guidance for managing swarm-size transitions to support operator workload in dynamic human--swarm systems. 
\end{abstract}


\maketitle              

\section{INTRODUCTION}
Decentralised robot swarms offer robustness and scalability in high-stakes environments; such safety-critical applications typically require human oversight for supervision, ethical judgement, and high-level control~\cite{lyons2025examining}. Human–Swarm Interaction (HSI) is the discipline of designing interfaces and interaction mechanisms that best utilise the human within the human–swarm team. This often involves balancing operator workload: the human must be engaged without being overloaded. Optimising workload in human–swarm teams therefore requires a combination of psychological insight and interface design, influencing how the swarm is controlled, commanded, and visualised~\cite{A2015Human}.

Workload history is the concept whereby the effects of a recent period of high demand persist after workload is reduced, often impacting performance~\cite{workloadHistory}. These hysteresis effects, or those where the relationship is dependant on the history or direction of the change, have been observed for mental workload~\cite{jansen2016hysteresis}. Such effects may arise in swarm operations, where the number of agents an operator must monitor can change dynamically. Some tasks benefit from gradual transitions in workload rather than abrupt shifts, although for drone control, varying workload over time has been shown to outperform maintaining a constant level, potentially by regulating limited cognitive resources~\cite{devlin2020transitions}. The relationship between swarm size and cognitive load more generally depends on interaction modality and task structure, with plateaus or inflection points often emerging. Sudden changes that induce shock or surprise are well studied in aerospace contexts, where pilots are trained to manage them~\cite{startle2}; in swarm settings, such events may be more frequent and less predictable due to the difficulty of monitoring many distributed assets simultaneously.

We propose the following hypotheses: \textbf{H1} --- Following a period of increased cognitive demand (larger swarms), operators' workload remains higher when the swarm size is reduced, compared to episodes at the same swarm size without a prior overload; \textbf{H2} --- For a given swarm size, operators retain lower workload when that swarm size is reached via gradual increases rather than via a single abrupt increase; \textbf{H3} --- Larger jumps in swarm size produce larger deviations from the no-change baseline than smaller jumps, with workload deviating more strongly following larger changes.

\section{BACKGROUND}
HSI often pursues ``scale invariance'', whereby an operator can effectively manage a swarm regardless of its size by abstracting away per-agent detail~\cite{Meyer2022A}. Closely related is the notion of ``fan-out'', which captures how many robots a human can reasonably supervise at a required level of performance~\cite{J2019Human}. Importantly, this limit is not fixed; it varies widely depending on task demands, interaction modality, and swarm behaviour, making it impractical to define a universal swarm size that can be managed~\cite{Chandarana2018Determining}. A range of interaction paradigms have been explored to support scalable HSI, most commonly through GUI-based control that allows operators to task individual robots or influence collective behaviour indirectly~\cite{A2013Human}. More abstract representations, such as heatmaps or aggregate visualisations, can reduce workload by suppressing low-level detail, but are not universally preferred; operators often still require access to individual agents for tasks such as fault diagnosis~\cite{Divband2021Designing}.

Central to human–swarm teaming is the management of mental workload. Overload can impair performance when operators are unable to keep up with task demands, while underload may be detrimental to engagement or vigilance~\cite{JA2023Can}. Mental workload is often assessed using subjective measures such as the NASA Task Load Index (TLX)~\cite{hart1988nasatlx}, which is widely accepted but necessarily retrospective and unsuitable for real-time adaptation. Consequently, physiological measures such as electroencephalography~\cite{Singh2025sUASInterfaces}, heart rate variability~\cite{Marois2024InterruptionsSurveillance}, and pupilometry have been explored in HSI~\cite{D2019Planetary}. The effect of swarm size on operator workload has been explored to a limited extent, but existing evidence suggests that larger swarms tend to increase perceived difficulty, arousal, and cognitive demand~\cite{Julian2023Effects}. As swarm size increases, operators are exposed to a greater number of low-level movements and interactions, which can elevate workload even in relatively simple monitoring tasks~\cite{CE2014Biologically}, with similar effects observed when managing multiple subswarms~\cite{J2024From}.

Most experimental studies assume a fixed swarm size for the duration of a task. In practice, however, swarm size is often dynamic: robots may fail, be redeployed, or join the mission as demands change, leading to abrupt workload fluctuations~\cite{ramchurn2016disaster}. Scalability is a defining property of swarms, but in HSI this scalability must extend to the human operator as well~\cite{B2013Scalable}. Prior work suggests that performance may improve with increasing swarm size up to a point before declining as operators fatigue over time~\cite{morrow2024evaluation}. While adaptive approaches have been proposed to monitor or respond to workload changes at runtime~\cite{abioye2024adaptivehumanswarminteractionbased}. The effect of these swarm-size changes on workload is still not well understood.


\section{METHODOLOGY}
\begin{figure}
\centering
\begin{minipage}[t]{0.54\linewidth}
    \centering
    \includegraphics[width=\linewidth]{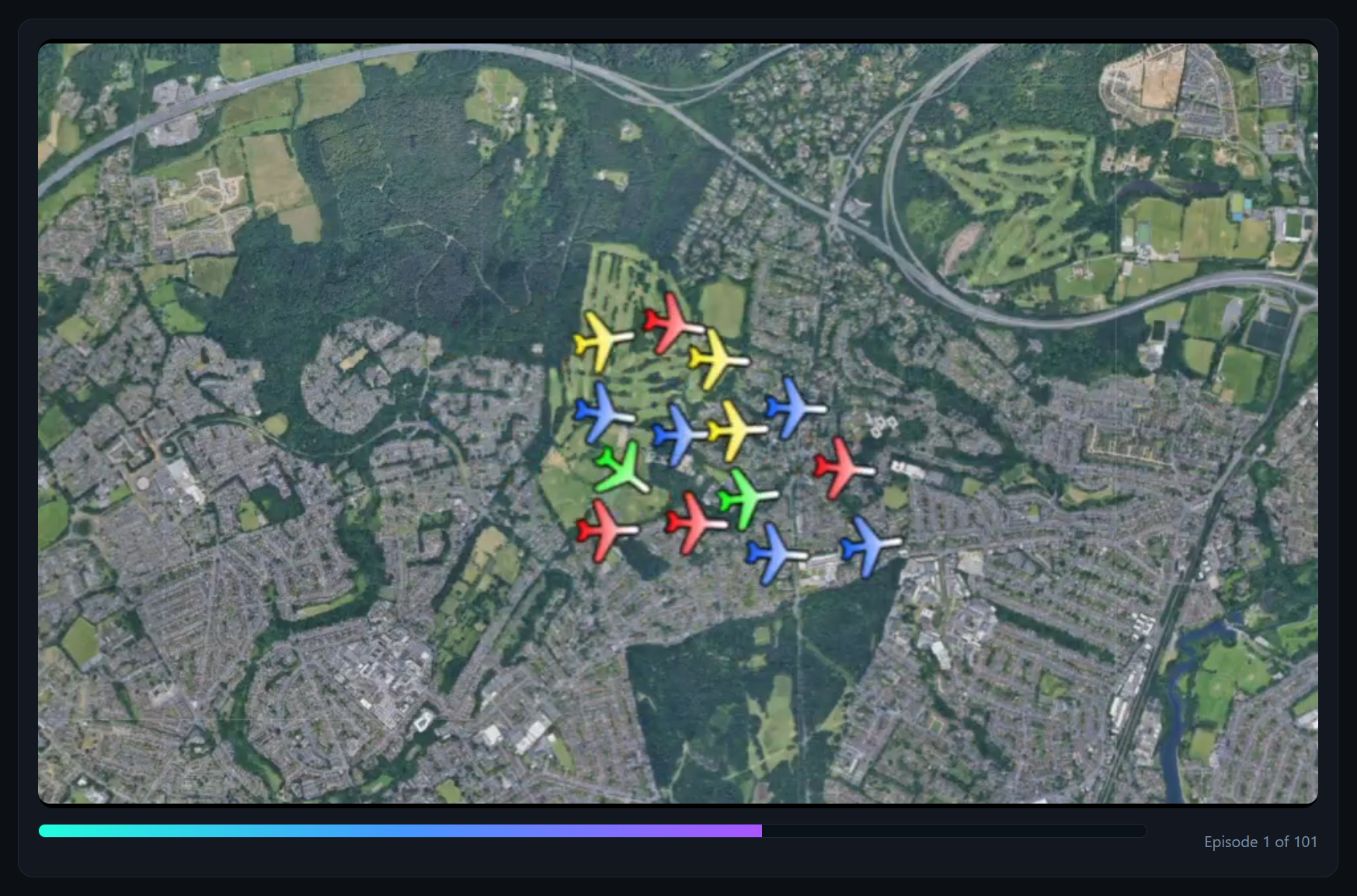}
    \captionof{figure}{Video shown to the user. Drones flock from left to right, and one disappears.}
    \label{fig:HARIS:Exp}
\end{minipage}\hfill
\begin{minipage}[t]{0.44\linewidth}
    \centering
    \includegraphics[width=\linewidth]{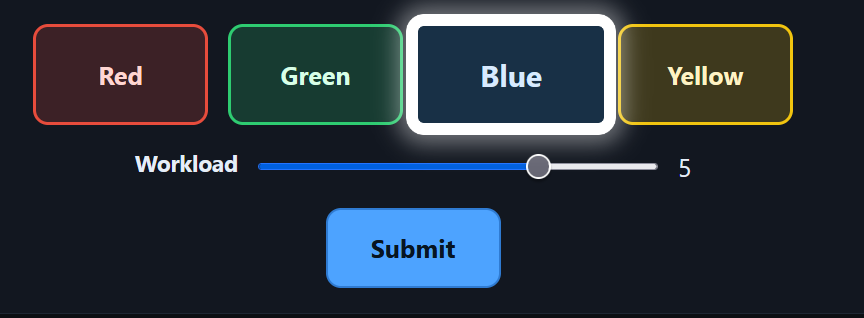}
    \captionof{figure}{Review panel. After the video ends, the user reports which colour drone disappeared and their workload.}
    \label{fig:HARIS:Rev}
\end{minipage}
\end{figure}

We design a simple experiment that measures an operator's ability to monitor a swarm for changes. A swarm of $N$ drones flocks across the map; partway through, one drone disappears, and the user must report the vanished drone's colour. This is not intended to model real drone failure (which could often be flagged automatically), rather serving as a measure for the operator's capacity to keep track of many entities within the swarm. We repeat this task across varying values of $N$, measuring both accuracy and self-reported mental workload. We define an episode $E(N)$ as a 7-second clip of $N$ drones flocking across the map using the boids algorithm~\cite{reynolds1987flocks}. Drones are divided between four colours (red, blue, green, yellow). At a random time between 2--4 seconds into the episode, one random drone disappears; at the end, the user reports its colour. Figures \ref{fig:HARIS:Exp} and \ref{fig:HARIS:Rev} shows an example episode. All episodes were recorded in the Human And Robot Interactive Swarm (HARIS) simulator~\cite{soorati2024enabling}. After each episode, participants also report their mental workload on a 1--7 scale.

Also of interest is the effect of the previous episode. In line with hypotheses \textbf{H1} and \textbf{H2}, we treat accuracy and subjective workload as being reported partly relative to the preceding swarm size. We describe an episode as $E(P,N)$, where 
$P$ and $N$ are the number of agents in the \textit{previous} and \textit{current} episodes respectively; The change size is $\Delta=N-P$ ($\Delta$ may be negative). We assume \textit{first-order dependence}: performance and workload in an episode depend on $N$ and the immediately prior $P$, but not on earlier episodes. Episodes are grouped into ``decks'' which are shown sequentially to participants; deck construction differs between studies, as described below.

\section{EXPERIMENT DESIGN}
We split the experiment into two studies, one testing \textbf{H1} and \textbf{H2} using small step changes in swarm size, and one testing \textbf{H3} using larger jumps to examine potential shock effects. Both studies were conducted on the same platform. Ethics approval was obtained through the University of Southampton ethics committee (ERGO-108904). Participants were recruited via Prolific, with inclusion criteria of age 18--65, residence in the UK or USA, and first-language English; colour-blindness was the only exclusion criterion. Participants completed the task on the Gorilla platform, first passing an Ishihara colour-vision test, then watching a 5-minute instructional video and completing a short practice tutorial. The tutorial also served as an attention check: participants had to score at least $3/7$ and provide varied workload reports to proceed. Participants then answered brief demographic questions, completed the experiment, and finished with a post-task survey. Our primary metrics were accuracy in identifying the disappeared drone's colour and self-reported workload after each episode. We additionally collected post-task subjective data to help contextualise the results.

\subsection{Study 1 -- Directional Effects of Small Swarm Size Changes}
Study 1 tests workload residue under decreasing swarm size (\textbf{H1}) and easing-in under increasing swarm size (\textbf{H2}). We used 9 swarm sizes ($N \in \{4,6,8,10,12,14,16,18,20\}$) with transitions $\Delta \in \{0,-2,+2\}$, each repeated $r=4$ times. To balance transitions, we generated episode decks as an Eulerian cycle over nodes $N$, where edges correspond to episodes $E(a,b)$. Using a modified Hierholzer algorithm\footnote{The original algorithm is deterministic. We choose a random start node and shuffle outgoing edges at each step, producing a different valid cycle per run.}, each possible transition appears exactly four times, excluding the initial ``warm-up'' episode. Two complementary decks were generated to reduce positional bias. The post-task survey focused on participants' perceptions of swarm size and size changes. Participants rated how larger swarms affected workload (5-point Likert), then provided open-ended reflections and Likert ratings on whether increases or decreases felt easier or harder. To capture emotional response, particularly surprise, we included a short subset of the PANAS-X questionnaire~\cite{watson1994panas}, consisting of 13 items spanning Attentiveness, Fatigue, Serenity, and Surprise, plus an ``other'' option, administered for both increase and decrease conditions. Finally, participants reported their task strategy and perceived relationships between swarm size, workload, and accuracy. We recruited 16 participants.

\subsection{Study 2 -- Directional Effects of Large Swarm Size Changes}
Study 2 followed the same structure as Study 1 but targeted larger shifts in swarm size to test \textbf{H3}. We balanced colour distributions to maintain engagement throughout each episode, then used 5 swarm sizes ($N \in \{8,12,16,20,24\}$) with transitions $\Delta \in \{0,\pm4,\pm8,\pm12,\pm16\}$ over $r=3$ repetitions. Four complementary decks were generated to reduce positional bias. The post-task survey additionally included the NASA Task Load Index (TLX)~\cite{hart1988nasatlx} to capture overall workload beyond per-episode ratings. We recruited 18 participants.

\section{RESULTS}
\label{sec:results}
We present the results of each study separately in the context of its focus.

\subsection{Study 1}
\begin{figure}[H]
    \centering
    \begin{minipage}{0.48\linewidth}
        \centering
        \includegraphics[width=\linewidth]{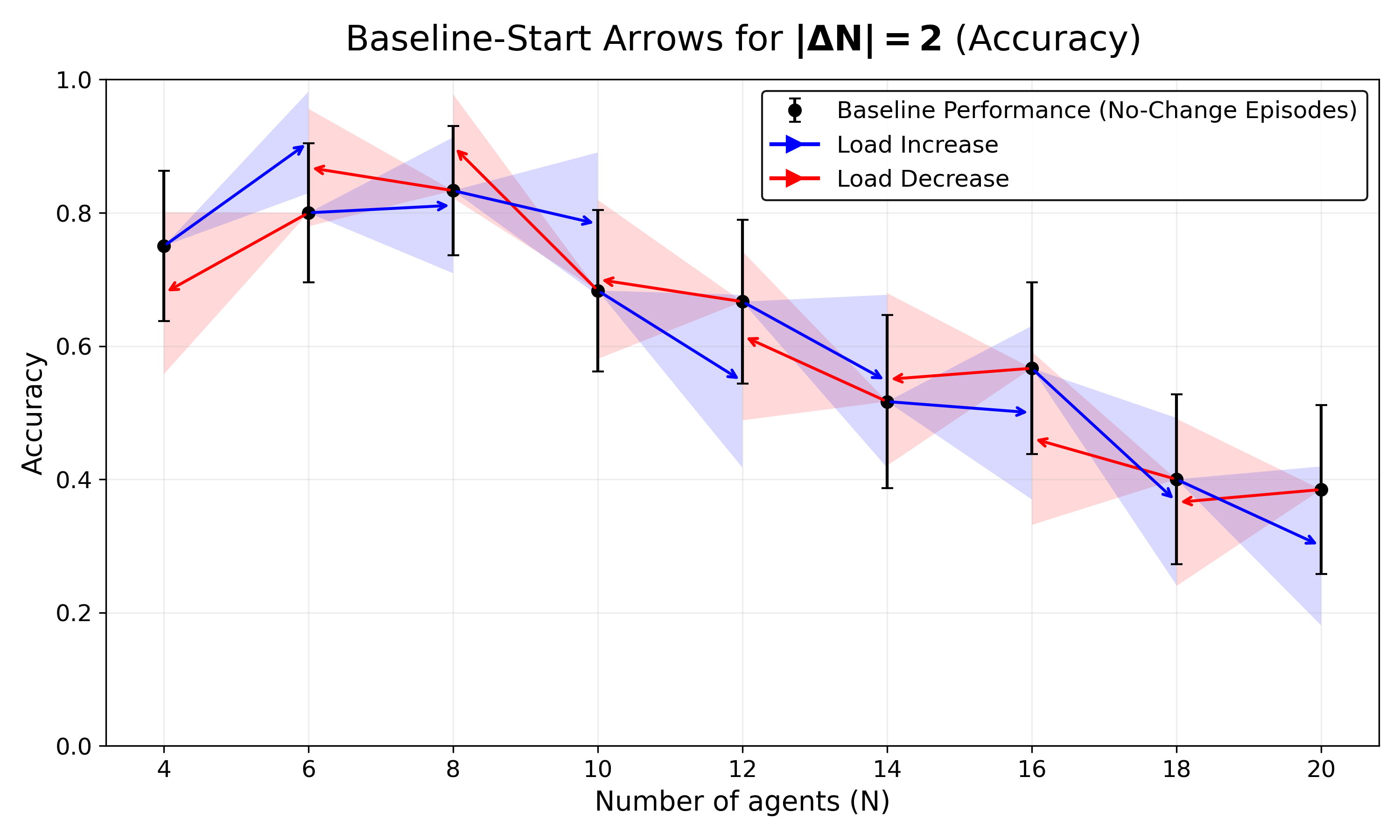}
    \end{minipage}\hfill
    \begin{minipage}{0.48\linewidth}
        \centering
        \includegraphics[width=\linewidth]{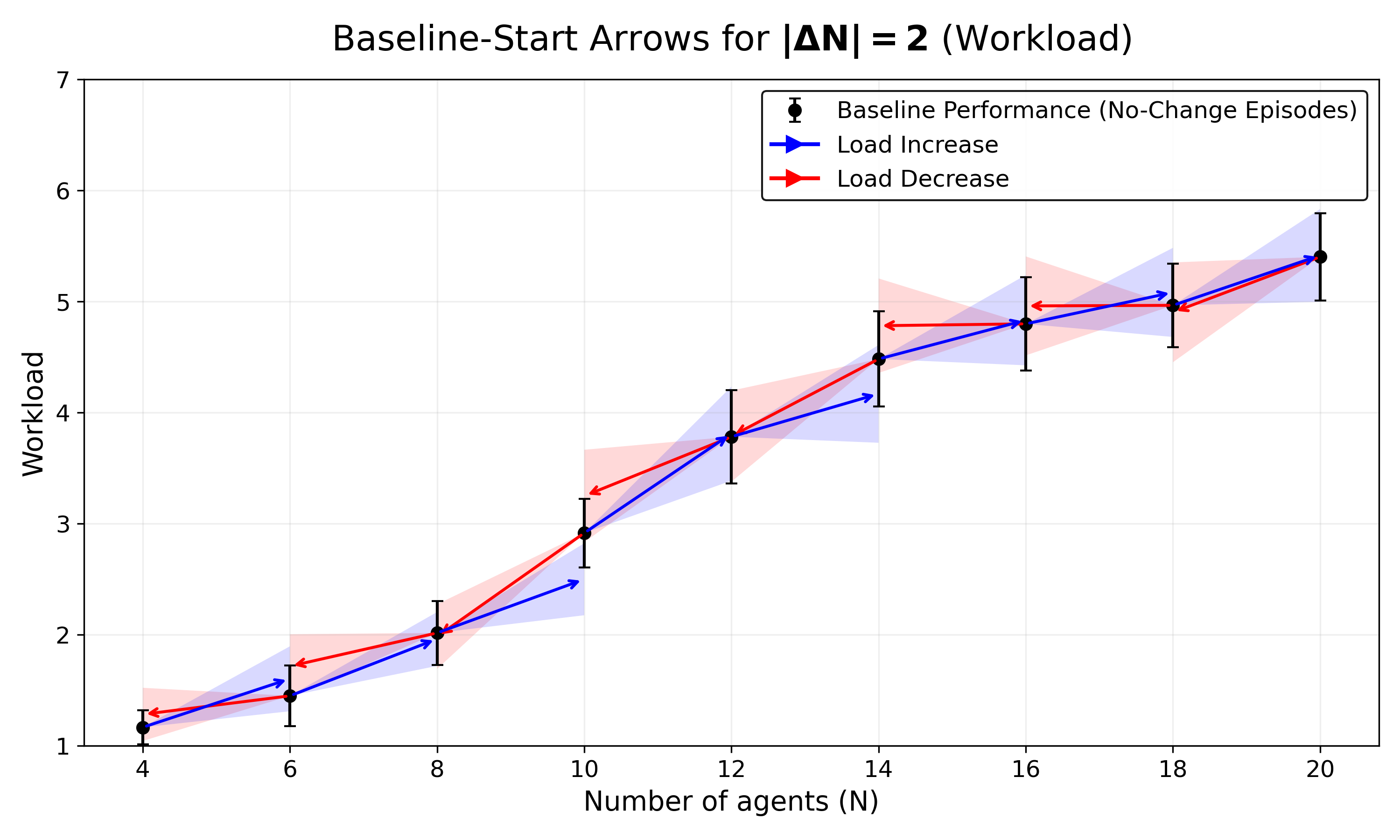}
    \end{minipage}
    \caption{Directional arrows showing changes vs baseline (black dots). E.g. A blue arrow from $8\rightarrow10$ agents represents $E(8,10)$. Shaded area around arrows and bars on baselines denote standard error of the mean}
    \label{fig:s1:del2}
\end{figure}

The results for small changes in swarm size were modest but promising. Shown in figure \ref{fig:s1:del2} is the baseline (all cases where the number of drones did not change), along with arrows showing the accuracy or workload resulting from each transition. Accuracy results are variable, with the direction of change having minimal impact on performance outside. Regarding workload, increases generally resulted in equal or lessened workload, while decreases resulted in equal or higher workload. This pattern is consistent with \textbf{H1} and \textbf{H2}: on the increase, operators appear to be ``eased in'', while on the decrease there exists a ``workload residue''.

\subsection{Study 2}
\begin{figure}[H]
    \centering
    \begin{tabular}{cc}
        \includegraphics[width=0.49\linewidth]{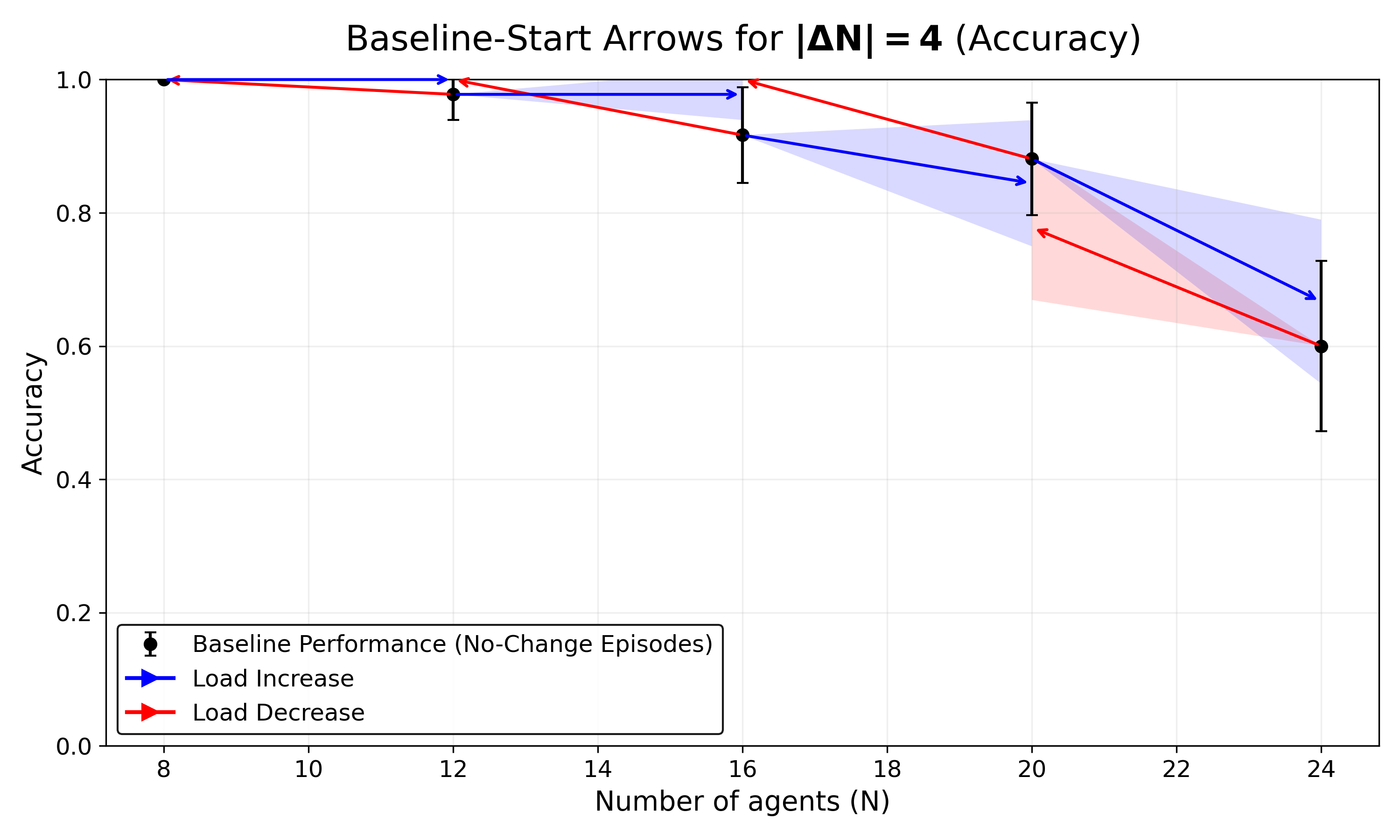} &
        \includegraphics[width=0.49\linewidth]{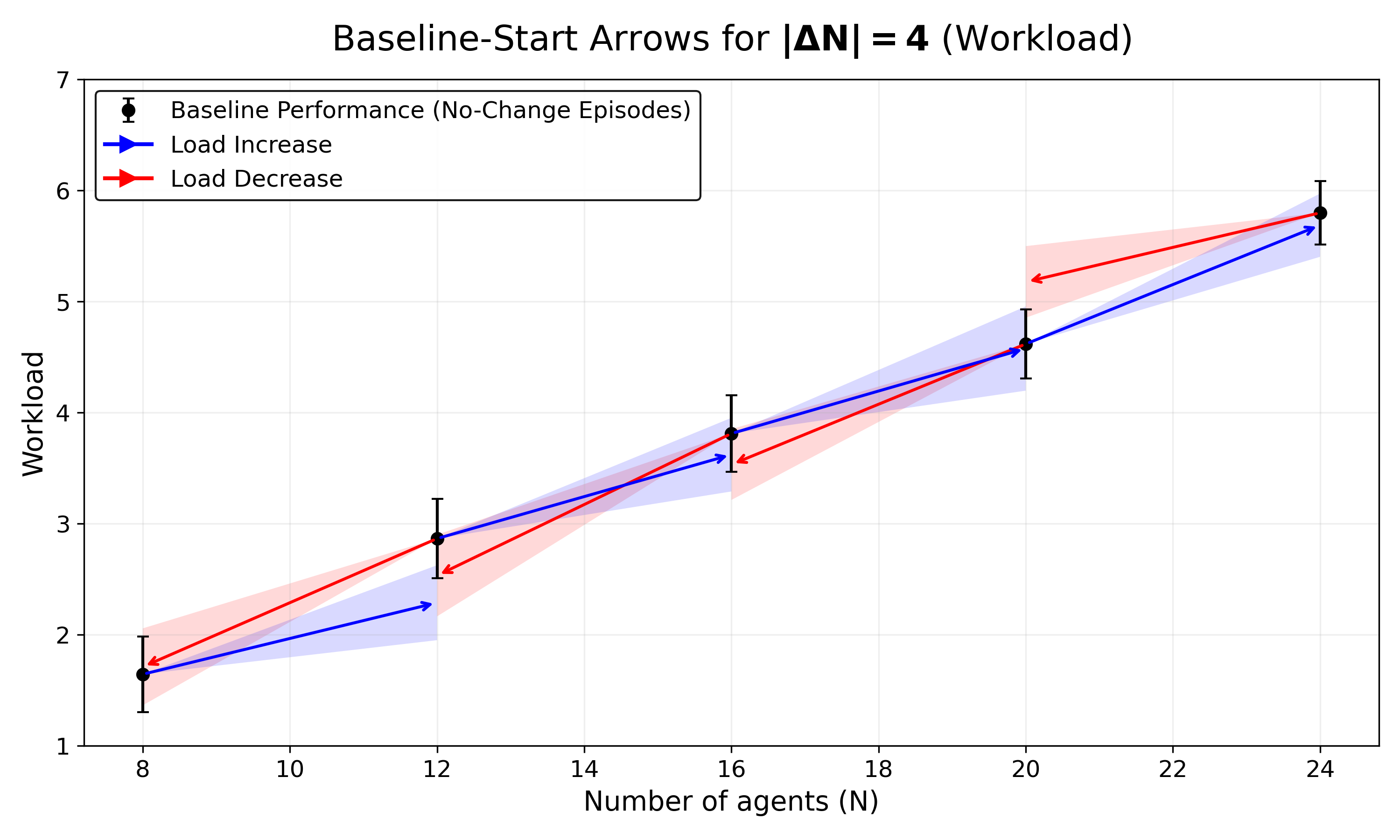} \\[-0.4em]

        \includegraphics[width=0.49\linewidth]{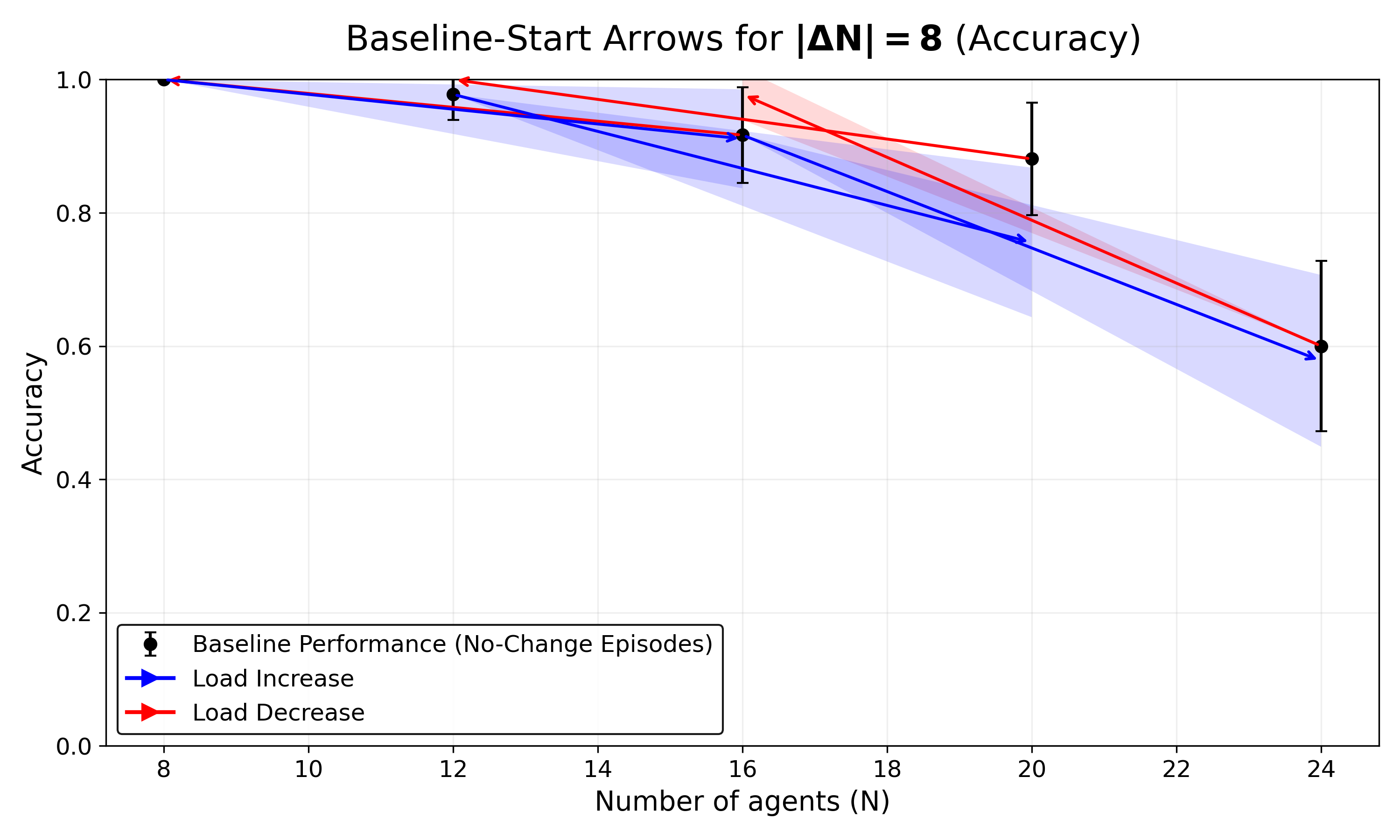} &
        \includegraphics[width=0.49\linewidth]{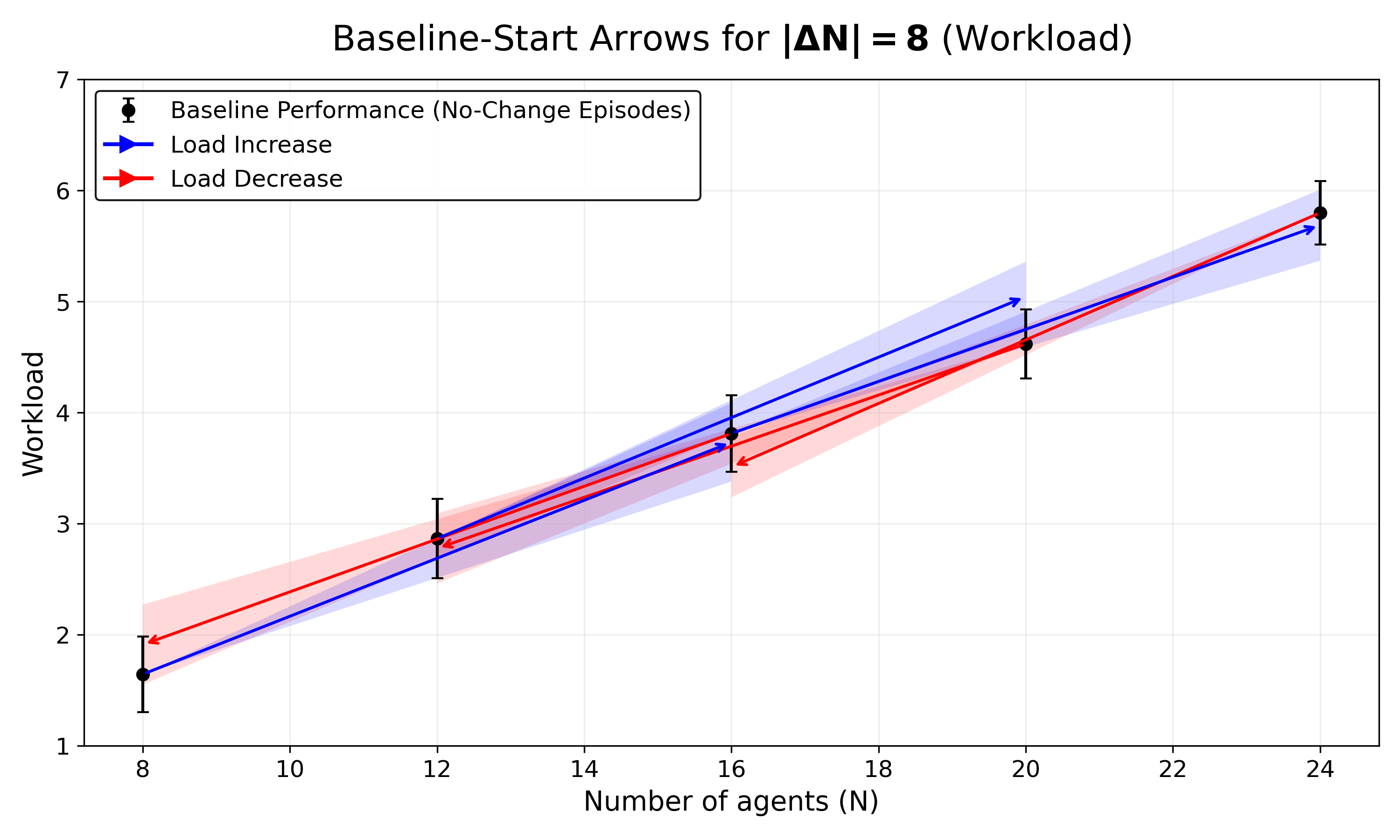} \\[-0.4em]

        \includegraphics[width=0.49\linewidth]{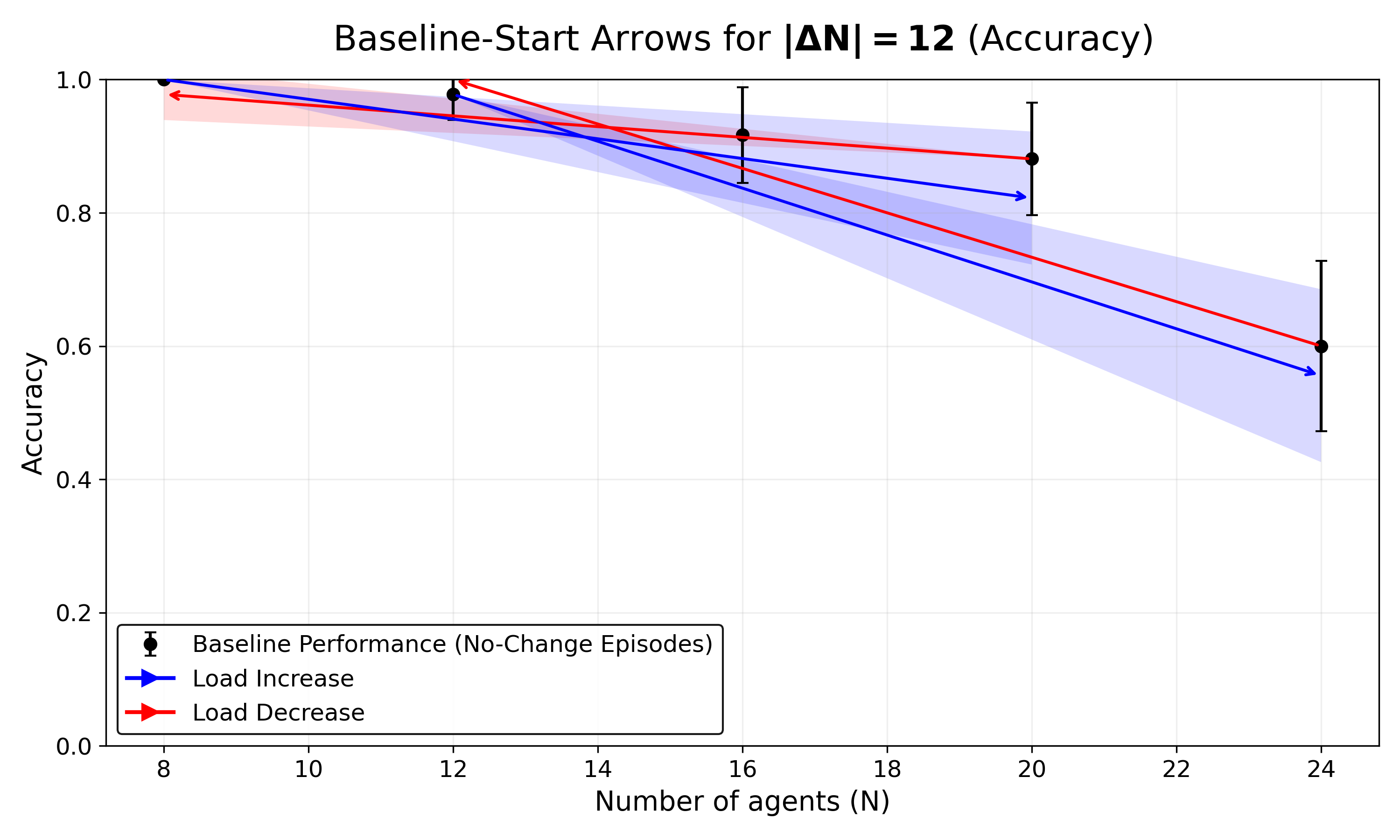} &
        \includegraphics[width=0.49\linewidth]{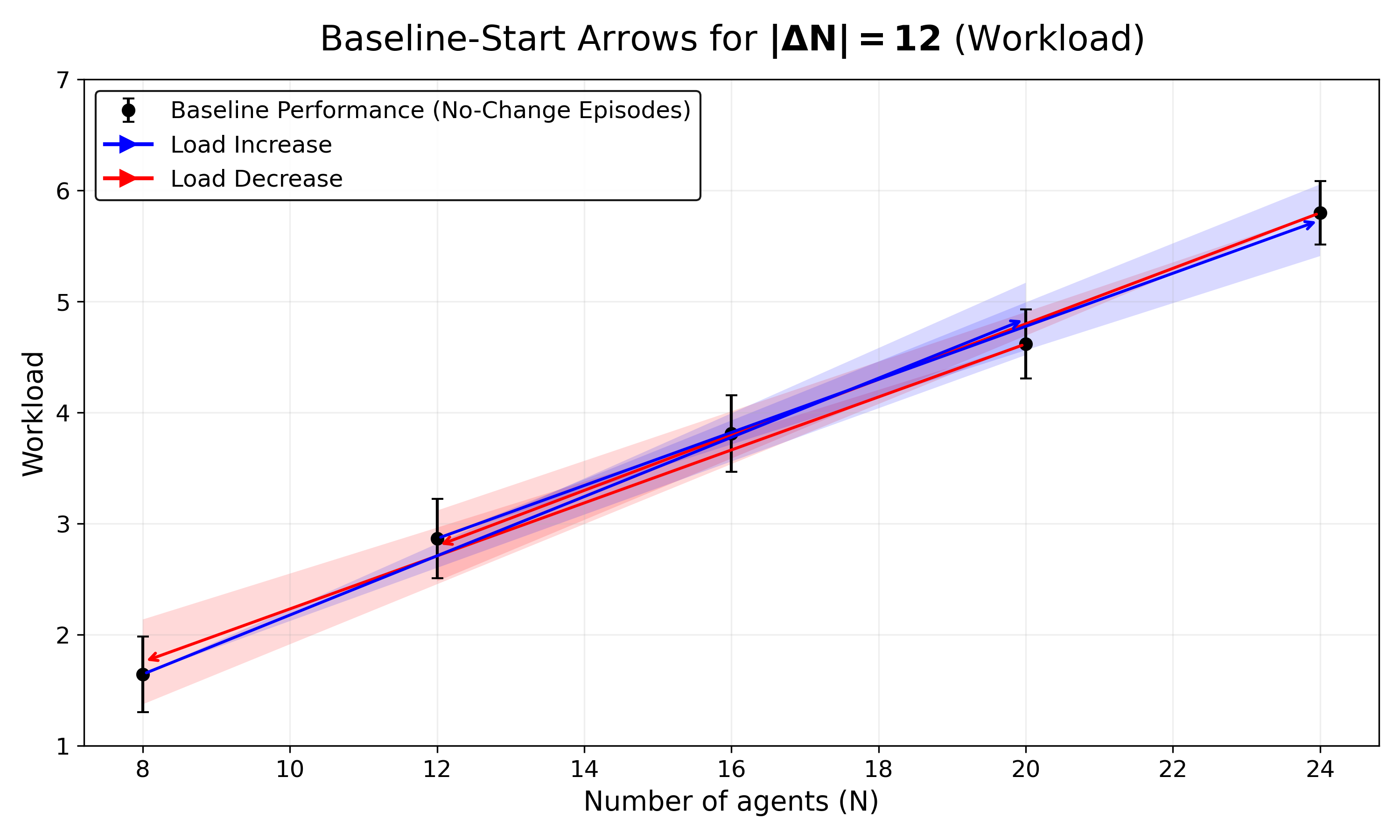}
    \end{tabular}
    \caption{Study 2 directional changes relative to baseline (black dots), grouped by swarm-size change magnitude. For each $\Delta$, accuracy (left) and workload (right) are shown. Shaded areas denote standard error of the mean.}
    \label{fig:study2_combined}
\end{figure}

Figure \ref{fig:study2_combined} shows arrow graphs for all adjacent ($\Delta=4$) changes. No clear pattern is seen for accuracy, however the workload results replicate those in Study 1: increases in the number of drones resulted in equal or lessened workload, while decreases tended to be associated with higher workload. This repetition strengthens the case for both \textbf{H2} and to a lesser extent \textbf{H1}.

Figure \ref{fig:study2_combined} shows arrow graphs for all medium ($\Delta=8$) changes; these changes are four-times larger than those in Study 1. A clearer pattern emerges at this jump size: increases in swarm size decrease accuracy, and decreases improve accuracy in all cases. Performance effects which were minimal at lower deltas may now be emerging; large increases in swarm size cause a drop in performance relative to baseline and large decreases cause a slight improvement. Although not directly related to our hypotheses, these results may be of some import. Workload differences begin collapsing at this magnitude of change; the deviations from baseline concordant with \textbf{H1} and \textbf{H2} are gone, supporting \textbf{H3} - that a large enough change ``shocks'' the user, invalidating the other two effects. Figure \ref{fig:study2_combined} shows similar results. The differences from baseline, especially in the workload case, get smaller as $\Delta$ increases, however the accuracy pattern seems to persist, and the workload differences are completely gone. This further supports that larger changes degrade the effects of \textbf{H1} and \textbf{H2}, supporting \textbf{H3}. Results for $\Delta=16$ are only a single pair of arrows so are not shown.

\subsection{Subjective Responses}

A consistent theme across Study~1 (where only small changes were tested) and Study~2 was that reductions in swarm size feel easier yet do not fully return workload to baseline. Several participants described immediate relief but also acknowledged lingering cognitive effort after a decrease. For example, P13 noted that it was ``a bit of a relief when the number dropped down'' because it gave their brain ``a second to relax after trying so hard on the larger number of drones,'' implying that the prior load persisted despite the decrease. P15 similarly stated that ``when the number of drones increased I felt more stressed and panicked; when the number decreased I felt relaxed,'' yet did not suggest a full reset to baseline. Another participant (P4) described feeling ``panicked'' by changes and ``relieved when I saw less drones,'' yet said the change ``was throwing me off,'' indicating lingering disruption. These accounts collectively support \textbf{H1}: small decreases ease workload but leave a residual cognitive burden. Participants' reflections on small increases were more muted; there were indications that gradual growth in swarm size was manageable, for example P1 explained that they ``always prepared for higher numbers'' and were ``pleasantly surprised'' when the increase was smaller, implying that anticipation and gradual changes allowed them to adapt. Despite this, participants were generally less supportive of \textbf{H2} overall, seldom mentioning small increases being preferable to larger ones.

Subjective responses most strongly endorsed the shock-severity hypothesis. Respondents frequently described large jumps in swarm size, especially increases, as surprising, disorienting, or panic-inducing. Participant~10 observed that when a trial with fewer drones was followed by one with many more, ``it increased my stress and decreased performance.'' Participants also reported slightly higher surprise for increases (mean $\approx 2.3$ on a 1--5 PANAS-X scale) than for decreases (mean $\approx 1.9$), indicating that sudden increases do indeed elicit a startle response. Together, these findings suggest that large, abrupt changes produce a qualitative shift in cognitive state, forcing operators to abandon systematic strategies and supporting \textbf{H3}.

\section{DISCUSSION}
Our first hypothesis (\textbf{H1}) concerned \textit{workload residue}, the idea that following overload, operators would perform worse or feel more burdened when the swarm size was subsequently reduced. Study~1 showed decreases in swarm size typically to resulted in similar or higher workload, although this started to break down at $\Delta=4$ and was not at all present for larger deltas. Subjective responses tended to support this idea. Overall, \textbf{H1} is fairly well supported at low deltas only.

Our second hypothesis (\textbf{H2}) concerned \textit{easing in}; that gradually increasing the swarm size up to $N$ agents would result in lower workload than an abrupt jump to the same size. Study~1 supported the easing-in effect, with increases generally yielding equal or lower workload than baseline. Study~2 replicated this for $\Delta = 4$, but it broke down for larger changes, which again is to be expected. Similarly to \textbf{H1}, \textbf{H2} is well supported, and our quantitative results suggested that it perhaps extends slightly further than \textbf{H1}, being present at $\Delta = 4$, whereas \textbf{H1} only held strongly at $\Delta=2$. This being said, the subjective experience of participants did not strongly support \textbf{H2}, although it is possible that our survey simply did not prompt the participants sufficiently well to reveal this.

Our third hypothesis (\textbf{H3}) concerned the \textit{shocking effect}. This proposed that larger jumps in swarm size would produce larger deviations in workload from the no-change baseline. This effect, if present, would compete with \textbf{H1} and \textbf{H2}; these two concepts may hold up to some jump size, after which the shocking effect overcomes the two small-change effects. While small jumps showed directional differences consistent with \textbf{H1} and \textbf{H2}, these differences \textit{collapsed} as deltas increased. Sufficiently large changes in swarm size caused any effects of \textbf{H1} and \textbf{H2} to be overcome by the magnitude of the shift, which we suspect is explainable through the mental resetting of the user. Subjective responses quite clearly supported startle and surprise effects, and multiple participants explicitly confirmed their own belief in \textbf{H3}. Overall, \textbf{H3} is supported - there exists a shocking effect which overcomes \textbf{H1} and \textbf{H2}.

\section{CONCLUSION}
Robotic swarms in real-world deployments are likely to be dynamic in size, and a single operator may have rapidly varying numbers of robots to monitor. We hypothesised that managing these size changes may be beneficial to human-swarm operator workload. \textbf{H1} - That there exists a ``workload residue'' when the swarm size slowly decreases, appears to suggest that although performance is mostly unaffected, operator workload does remain elevated during these slow reductions. \textbf{H2} - That a user could be ``eased-in'' slowly to a larger swarm size, holds for workload effects, although also has minimal performance impact. Finally \textbf{H3} - That larger changes ``shock'' the operator and invalidate the other two effects, seems to be correct. Our findings are promising, although they focus only on the operator's ability to monitor the swarm; in practice, most monitoring can be automated and instead it is tasking and attending to novel problems that the operator would spend more time on. Future work should explore these more complex human-swarm teaming missions. Regardless, it would behove swarm designers to be aware of changeable swarm sizes in real-world deployments, and consider balancing out increases and decreases where possible to ``ease-in'' the operators with small increases, and utilise larger changes to ``shock'' the user, overcoming the ``workload residue'' effect.

\section{ACKNOWLEDGMENTS}
This work was supported by the Trustworthy Autonomous Systems Hub [EP/V00784X/1]. The lead author was also supported by UK Research and Innovation [EP/S024298/1]

\bibliographystyle{apalike}
\bibliography{refs}

@article{A2013Human,
  author    = {Kolling, Andreas and Sycara, Katia and Nunnally, Sarah and Lewis, Michael},
  title     = {Human swarm interaction: An experimental study of two types of interaction with foraging swarms},
  year      = {2013},
  journal   = {Journal of Human-Robot Interaction},
  volume    = {2},
  number    = {2},
  pages     = {103--129},
  doi       = {10.5898/JHRI.2.2.Kolling}
}

@article{A2015Human,
  author    = {Kolling, Andreas and Walker, Phillip and Chakraborty, Nilanjan and Sycara, Katia and Lewis, Michael},
  title     = {Human interaction with robot swarms: A survey},
  year      = {2015},
  journal   = {IEEE Transactions on Human-Machine Systems},
  volume    = {46},
  number    = {1},
  pages     = {9--26},
  doi       = {10.1109/THMS.2015.2480801}
}

@misc{abioye2024adaptivehumanswarminteractionbased,
  title={Adaptive Human-Swarm Interaction based on Workload Measurement using Functional Near-Infrared Spectroscopy},
  author={Ayodeji O. Abioye and Aleksandra Landowska and William Hunt and Horia Maior and Sarvapali D. Ramchurn and Mohammad Naiseh and Alec Banks and Mohammad D. Soorati},
  year={2024},
  eprint={2405.07834},
  archivePrefix={arXiv},
  primaryClass={cs.RO},
  url={https://arxiv.org/abs/2405.07834}
}

@inproceedings{B2013Scalable,
  author    = {Pendleton, Brian and Goodrich, Michael},
  title     = {Scalable human interaction with robotic swarms},
  year      = {2013},
  doi       = {10.2514/6.2013-4731},
  booktitle = {AIAA Infotech@Aerospace (I@A) Conference}
}

@inproceedings{CE2014Biologically,
  author    = {Harriott, Caroline E. and Seiffert, Adriane E. and Hayes, Shanna T. and Adams, Julie A.},
  title     = {Biologically-inspired human-swarm interaction metrics},
  year      = {2014},
  doi       = {10.1177/1541931214581307},
  booktitle = {Proceedings of the Human Factors and Ergonomics Society Annual Meeting},
  journal   = {Proceedings of the Human Factors and Ergonomics Society Annual Meeting},
  volume    = {58},
  number    = {1},
  pages     = {1471--1475}
}

@inproceedings{Chandarana2018Determining,
  author = {M. Chandarana and M. Lewis and K. Sycara and S. Scherer},
  title = {Determining Effective Swarm Sizes for Multi-Job Type Missions},
  year = {2018},
  doi = {10.1109/IROS.2018.8593919},
  booktitle = {2018 IEEE/RSJ International Conference on Intelligent Robots and Systems (IROS)},
  pages = {4848--4853}
}

@article{D2019Planetary,
  title={Planetary exploration with robot teams: Implementing higher autonomy with swarm intelligence},
  author={St-Onge, David and Kaufmann, Marcel and Panerati, Jacopo and Ramtoula, Benjamin and Cao, Yanjun and Coffey, Emily BJ and Beltrame, Giovanni},
  journal={IEEE Robotics \& Automation Magazine},
  volume={27},
  number={2},
  pages={159--168},
  year={2019},
  publisher={IEEE}
}

@article{devlin2020transitions,
  title={Transitions between low and high levels of mental workload can improve multitasking performance},
  author={Devlin, Shannon Patricia and Moacdieh, Nadine Marie and Wickens, Christopher D and Riggs, Sara Lu},
  journal={IISE transactions on occupational ergonomics and human factors},
  volume={8},
  number={2},
  pages={72--87},
  year={2020},
  publisher={Taylor \& Francis}
}

@article{Divband2021Designing,
  title={Designing a user-centered interaction interface for human--swarm teaming},
  author={Divband Soorati, Mohammad and Clark, Jediah and Ghofrani, Javad and Tarapore, Danesh and Ramchurn, Sarvapali D},
  journal={Drones},
  volume={5},
  number={4},
  pages={131},
  year={2021},
  publisher={Multidisciplinary Digital Publishing Institute}
}

@incollection{hart1988nasatlx,
  author    = {Hart, Sandra G. and Staveland, Lowell E.},
  year      = {1988},
  title     = {Development of NASA-TLX (Task Load Index): Results of Empirical and Theoretical Research},
  booktitle = {Human Mental Workload},
  editor    = {Hancock, Peter A. and Meshkati, Najmedin},
  pages     = {139--183},
  publisher = {North-Holland}
}

@inproceedings{J2019Human,
  author    = {Humann, James and Pollard, Kimberly A.},
  title     = {Human factors in the scalability of multirobot operation: A review and simulation},
  year      = {2019},
  booktitle = {2019 IEEE International Conference on Systems, Man and Cybernetics (SMC)},
  pages     = {700--707},
  doi       = {10.1109/SMC.2019.8913876}
}

@inproceedings{morrow2024evaluation,
  title={Evaluation of the Human-Robot-Interaction Dynamic Under Mental Fatigue Constraints In Search and Rescue Operations},
  author={Morrow, Jordan and Zawodniok, Maciej},
  booktitle={2024 International Conference on Information and Communication Technologies for Disaster Management (ICT-DM)},
  pages={1--7},
  year={2024},
  organization={IEEE}
}

@inproceedings{J2024From,
  author    = {Kaduk, Julian and Cavdan, Müge and Drewing, Knut and Vatakis, Argiro and Hamann, Heiko},
  title     = {From one to many: How active robot swarm sizes influence human cognitive processes},
  year      = {2024},
  booktitle = {2024 33rd IEEE International Conference on Robot and Human Interactive Communication (RO-MAN)}
}

@article{JA2023Can,
  author = {Adams, JA and Hamell, J and Walker, P},
  title = {Can a single human supervise a swarm of 100 heterogeneous robots?},
  year = {2023},
  journal = {Field Robotics}
}

@article{jansen2016hysteresis,
  title={Hysteresis in mental workload and task performance: the influence of demand transitions and task prioritization},
  author={Jansen, Reinier J and Sawyer, Ben D and Van Egmond, Ren{\'e} and De Ridder, Huib and Hancock, Peter A},
  journal={Human factors},
  volume={58},
  number={8},
  pages={1143--1157},
  year={2016},
  publisher={SAGE Publications Sage CA: Los Angeles, CA}
}

@inproceedings{Julian2023Effects,
  author = {Kaduk, Julian and Cavdan, Müge and Drewing, Knut and Vatakis, Argiro and Hamann, Heiko},
  title = {Effects of Human-Swarm Interaction on Subjective Time Perception: Swarm Size and Speed},
  year = {2023},
  booktitle = {Proceedings of the 2023 ACM/IEEE International Conference on Human-Robot Interaction},
  pages = {456--465}
}

@article{lyons2025examining,
  title={Examining the human-centred challenges of human--swarm interaction},
  author={Lyons, Joseph B and Capiola, August and Adams, Julie A and Mator, Janine D and Cherry, Erin and Barrera, Kristen},
  journal={Philosophical Transactions A},
  volume={383},
  number={2289},
  pages={20240140},
  year={2025},
  publisher={The Royal Society}
}

@inproceedings{Meyer2022A,
  author = {J. Meyer and A. Pinosky and T. Trzpit and E. Colgate and T. D. Murphey},
  title = {A Game Benchmark for Real-Time Human-Swarm Control},
  year = {2022},
  doi = {10.1109/CASE49997.2022.9926423},
  booktitle = {2022 IEEE 18th International Conference on Automation Science and Engineering (CASE)},
  pages = {743--750}
}

@article{ramchurn2016disaster,
  title={A disaster response system based on human-agent collectives},
  author={Ramchurn, Sarvapali D and Huynh, Trung Dong and Wu, Feng and Ikuno, Yukki and Flann, Jack and Moreau, Luc and Fischer, Joel E and Jiang, Wenchao and Rodden, Tom and Simpson, Edwin and others},
  journal={Journal of Artificial Intelligence Research},
  volume={57},
  pages={661--708},
  year={2016}
}

@inproceedings{reynolds1987flocks,
  title={Flocks, herds and schools: A distributed behavioral model},
  author={Reynolds, Craig W},
  booktitle={Proceedings of the 14th annual conference on Computer graphics and interactive techniques},
  pages={25--34},
  year={1987}
}

@incollection{soorati2024enabling,
  title={Enabling trustworthiness in human-swarm systems through a digital twin},
  author={Soorati, Mohammad D and Naiseh, Mohammad and Hunt, William and Parnell, Katie and Clark, Jediah and Ramchurn, Sarvapali D},
  booktitle={Putting AI in the Critical Loop},
  pages={93--125},
  year={2024},
  publisher={Elsevier}
}

@article{startle2,
  author = {Alexandre Duchevet and Jean-Paul Imbert and Jérémie Garcia and Benoît Lamirault and Mickaël Causse},
  title ={Investigating the Independent and Combined Effects of Startle and Surprise in a Simulated Flight Task},
  journal = {Human Factors},
  volume = {67},
  number = {11},
  pages = {1170-1187},
  year = {2025},
  doi = {10.1177/00187208251342100},
  note ={PMID: 40373188}
}

@article{watson1994panas,
  title={The PANAS-X: Manual for the positive and negative affect schedule-expanded form},
  author={Watson, David and Clark, Lee Anna},
  year={1994},
  publisher={University of Iowa}
}

@article{workloadHistory,
  author = {Fuenzalida, Eugenia},
  year = {2007},
  month = {05},
  pages = {277-91},
  title = {Effect of Workload History on Task Performance},
  volume = {49},
  journal = {Human factors},
  doi = {10.1518/001872007X312496}
}

@inproceedings{Singh2025sUASInterfaces,
  title        = {Optimizing Human-Machine Interfaces for Neuroergonomics: Cognitive Workload and Performance in sUAS Operations},
  author       = {Singh, Suvipra},
  booktitle    = {Human-Computer Interaction \& Emerging Technologies (AHFE Conference Proceedings)},
  year         = {2025},
  doi          = {10.54941/ahfe1006227},
  url          = {https://openaccess.cms-conferences.org/publications/book/978-1-964867-71-7/article/978-1-964867-71-7_9}
}

@inproceedings{Marois2024InterruptionsSurveillance,
  title        = {Using Cardiac and Electrodermal Activity as Cognitive Markers for Interruptions and Distraction in a Surveillance Simulation},
  author       = {Marois, Alexandre and Mouratille, Damien and Pratviel, Yvan and Chamberland, Cindy and Tremblay, S{\'e}bastien},
  booktitle    = {Neuroergonomics and Cognitive Engineering (AHFE Conference Proceedings)},
  year         = {2024},
  doi          = {10.54941/ahfe1004733},
  url          = {https://openaccess.cms-conferences.org/publications/book/978-1-964867-02-1/article/978-1-964867-02-1_0}
}

\end{document}